\documentclass[10pt,runningheads]{llncs}

\usepackage{booktabs}  
\usepackage{multirow}
\usepackage{bm}
\usepackage{array}
\usepackage{graphicx}
\usepackage{epstopdf}
\usepackage{subfigure}
\usepackage{amssymb}
\usepackage{amsmath}
\usepackage{makecell}
\usepackage[pagebackref=false,breaklinks=true,letterpaper=true,colorlinks,bookmarks=false]{hyperref}
\usepackage{cite}
\graphicspath{{figs/}}

\usepackage[draft,authormarkuptext=none,addedmarkup=none]{changes}
\colorlet{Changes@Color}{blue}

\emergencystretch=\hsize
\begin{document}

\title{Residual Network based Aggregation Model for Skin Lesion Classification }

\titlerunning{Aggregate ResNet For Skin Lesion Classification}

\author{Yongsheng Pan$^{1,2}$, 
    Yong Xia$^{1*}$ (yxia@nwpu.edu.cn)}

\institute{
$^1$ School of Computer Science and Engineering, Northwestern Polytechnical University, Xi'an 710072, China\\
$^2$ Department of Radiology and BRIC, University of North Carolina at Chapel Hill, North Carolina 27599, U.S.A. 
}

\maketitle

\begin{abstract}

We recognize that the skin lesion diagnosis is an essential and challenging sub-task in Image classification, in which the Fisher vector (FV) encoding algorithm and deep convolutional neural network (DCNN) are two of the most successful techniques. Since the joint use of FV and DCNN has demonstrated proven success, the joint techniques could have discriminatory power on skin lesion diagnosis as well. To this hypothesis, we propose the aggregation algorithm for skin lesion diagnosis that utilize the residual network to extract the local features and the Fisher vector method to aggregate the local features to image-level representation. We applied our algorithm on the International Skin Imaging Collaboration 2018 (ISIC2018) challenge and only focus on the third task, i.e., the disease classification.  

\end{abstract}

\section{Introduction}
In recent years, deep learning techniques have been widely acknowledged as the most powerful tool for image classification, since various DCNNs, such as VggNet \cite{chatfield2014return,simonyan2014very} and ResNet \cite{he2016deep}, have won the ImageNet Challenge in recent years. However, it also has been widely criticized that DCNNs may suffer from over-fitting when the training dataset is not large enough \cite{szegedy2013deep}.Although a pre-trained DCNN model can transfer the image representation ability learned from large-scale datasets, such as ImageNet, to the generic visual recognition tasks \cite{lin2017bilinear,girshick2014rich,ge2017borrowing,li2016compact,gao2016compact,kong2017low,yao2016coarse}, the rigid architectures of DCNNs limit the ability in dealing with images where objects have large variation in shape, size and clutter. 

To overcome this drawback, the feature maps learned by a pre-trained DCNN are viewed as local descriptors and aggregated to a high-level image representation by using the FV encoding method \cite{perronnin2007fisher,sanchez2013image}. Combining with FV has become the most competitive strategy to boost the performance of DCNN in image classification tasks \cite{palasek2017discriminative,cimpoi2016deep,tang2016deep,liu2017compositional}. 

In this paper, we introduce the DCNN-FV approach for skin lesion diagnosis. We use the pre-trained Residual Network, e.g. ResNet50, ResNet101 to extract multi-scale features for each image and aggregate them to image-level representation for classification. We applied our algorithm on the dataset published by the International Skin Imaging Collaboration 2018 (ISIC2018) and only focus on the third task, i.e., the disease classification. We trained and test our algorithm on the training data and validation data, and report the results achieved by the submission website. 

\section{Theory of Aggregation Model}

Regarding image classification, each image consists of class-relevant foreground and class-irrelevant background, where only the foreground is related to the classification task and has the discriminatory power.
Let the local descriptors extracted in foreground and background of an image follow the distribution $ q $ and $ r $, respectively. Thus, the distribution of all local descriptors in this image is 
\begin{equation} 
p(x)=wq(x)+(1-w)r(x),
\label{eq10}
\end{equation}
where $ w $ is the proportion of $ x $ being extracted from the class-relevant foreground.The generation process can be written as
\begin{equation} 
G_\lambda^X=\mathbb{E}_{x\sim q}\log{⁡u_\lambda(x_t )}={\nabla_\lambda}\int_x q(x)\log{⁡u_\lambda(x)}dx. 
\label{eq11}
\end{equation}

Since $ u_\lambda $ is estimated according to the maximum likelihood principle (MLP) without differentiating the class-relevant descriptors and class-irrelevant descriptors in each image, there exists a bias $ \epsilon_\lambda $ between the real distribution $ u_\lambda $ and the estimated distribution $ \hat{u}_\lambda $ as
\begin{equation} 
\hat{u}_λ=\alpha u_\lambda+(1-\alpha) \epsilon_\lambda,
\label{eq15}
\end{equation}
where $ \alpha $ is the proportion. Therefore, the estimated gradient has the insufficient form as
\begin{equation}
\hat{G}_\lambda^X={\nabla_\lambda}\int_x p(x)\log{\hat{u}_\lambda(x)}dx.
\label{eq16}
\end{equation}

On the other hand, when applying \eqref{eq10} to \eqref{eq11}, we have
\begin{equation} 
\begin{matrix}
\hat{G}_\lambda^X=&w{\nabla_\lambda}\int_x q(x)\log{\hat{u}_\lambda(x)} dx+ &\\
&(1-w){\nabla_\lambda}\int_x r(x)\log{\hat{u}_\lambda(x)} dx.&
\label{eq12}
\end{matrix}
\end{equation}
To eliminate the impact of $ r(x) $ to the class specialty of $ G_\lambda^X $, we need 
\begin{equation} 
(1-w){\nabla_\lambda}\int_x r(x)\log{\hat{u}_\lambda(x)}dx=0,
\label{eq13}
\end{equation}
where $ r $ and $ w $ are fixed when the image contents are constant. Once a suitable $ u_\lambda $ is found that satisfies \eqref{eq13}, we can simplify the generative model FV given in \eqref{eq12} as follows
\begin{equation} 
G_\lambda^X=w{\nabla_\lambda}\int_x q(x)\log{u_\lambda(x)}dx.
\label{eq14}
\end{equation}
This means that the generative model focuses on the class-relevant foreground. However, if we have no intuition to apply the bounding boxes or do segmentation, we can hypothesize \eqref{eq13} is satisfied and let $ G_\lambda^X = \hat{G}_\lambda^X$.

\section{Experiments}

\subsection{Materials}
Our data was extracted from the “ISIC 2018: Skin Lesion Analysis Towards Melanoma Detection” grand challenge datasets \cite{ISIC2018_1,ISIC2018_2}. This dataset consists of 11720 images in seven skin lesion categories. The distribution is very imbalanced and most of the samples are Melanoma. A split is provided that 10015, 173 and 1512 images for training, validation and testing. While doing the experiments, we only know the labels of training images and can acquire the scores of validation. 

\subsection{Implementation}
Our algorithm consists of two stages. In the first stage, we fine tuned the ResNet-50 15 epochs for local feature extraction. We first randomly cropped images with factors from 0.25 to 1.00 for data augmentation, and resized them to 224*224. The learning rate is 1e-3 for the first 10 epochs and 1e-4 for the next 5 epochs. In the second stage, we rescaled images with factors $ 2^s,s=-3,-2.5,\cdots,1 $. Then, these images were inputted to, which was pre-trained on ImageNet dataset and fine-tuned on the training data. We use the outputs of the layer 'res5c\_branch2a' of ResNet-50/101/152 \cite{he2016deep} as local descriptors whose dimension is 512. On each experimental trail, we trained a codebook of 64 Gaussian components with descriptors sampled from no more than 1000 images and Encoded the local descriptors of each image to a Fisher vector. The classification accuracy was measured by the balanced accuracy over classes (BAC).

\subsection{Results}
We reported the results of three algorithm, i.e., 1) extract feature via Fisher Vector with pre-trained ResNet without fine-tuning, 2) fine-tune pre-trained ResNet, 3) extract feature via Fisher Vector with fine-tuned ResNet. We mainly contained three backbone, i.e., the ResNet-50/101/152. All trails are trained on the public training images and test scores on the validation are listed on Table 1. The results show that the aggregation model with fine-tuned ResNet reach the best performance.

{
\setcellgapes{1.8pt}

\begin{table}[tp]
\setlength{\belowcaptionskip}{-0pt}
\setlength{\abovecaptionskip}{-1pt}
\setlength\abovedisplayskip{-1pt}
\setlength\belowdisplayskip{-1pt}
\renewcommand\arraystretch{1.0}
\centering
\caption{Performance of three methods on validation data.}

\begin{tabular*}{0.80\textwidth}{@{\extracolsep{\fill}}c ccccc}
\toprule
\multicolumn{1}{l}{Method}
 &FV+SVM & fine tuning & ftFV+SVM
\\
\hline
\multicolumn{1}{l}{ResNet-50}	   
&0.785  &0.824  &0.904  
\\ 

\multicolumn{1}{l}{ResNet-101}	
&0.749  &0.877  &0.884 
\\
\multicolumn{1}{l}{ResNet-152}	
&0.781  &0.822  &0.850   	
\\
\bottomrule
\end{tabular*} 
\label{tab_results}
\end{table}}

\subsubsection{Acknowledgment}
This research was supported in part by the National Natural Science Foundation of China under Grants 61771397 and 61471297, in part by Innovation Foundation for Doctor Dissertation of NPU under Grants CX201835.

\bibliographystyle{splncs}
\bibliography{References}

\end{document}